\documentclass[conference]{IEEEtran}
\IEEEoverridecommandlockouts

\usepackage{graphics}
\usepackage{graphicx}
\usepackage{amsmath}
\usepackage{amssymb}
\usepackage{amsfonts}
\usepackage{booktabs}
\usepackage{multirow}
\usepackage[table]{xcolor}
\usepackage{array}
\usepackage{makecell}
\usepackage{cite}
\usepackage{url}
\usepackage{times}
\usepackage{float} 

\usepackage[format=plain,labelfont=bf,textfont=normal]{caption}
\usepackage{subcaption}

\title{Intelligent Power Grid Design Review via Active Perception-Enabled Multimodal Large Language Models}

\author{
    \IEEEauthorblockN{
        Taoliang Tan\IEEEauthorrefmark{1},
        Chengwei Ma\IEEEauthorrefmark{2},
        Zhen Tian\IEEEauthorrefmark{2},
        Zhao Lin\IEEEauthorrefmark{2},
        Dongdong Li\IEEEauthorrefmark{2},
        Si Shi\IEEEauthorrefmark{2}\IEEEauthorrefmark{3}
    }
    \IEEEauthorblockA{\IEEEauthorrefmark{1}Yangjiang Yangxi Power Supply Bureau, Guangdong Power Grid Co., Ltd., Yangjiang, China}
    \IEEEauthorblockA{\IEEEauthorrefmark{2}Guangdong Laboratory of Artificial Intelligence and Digital Economy (SZ), Shenzhen, China}
    \IEEEauthorblockA{\IEEEauthorrefmark{3}Corresponding author: \texttt{shisi@gml.ac.cn}}
}

\begin{document}
\maketitle
\thispagestyle{empty}
\pagestyle{empty}
\renewcommand\IEEEkeywordsname{Index Terms}

\begin{abstract}
The intelligent review of power grid engineering design drawings is crucial for power system safety. However, current automated systems struggle with ultra-high-resolution drawings due to high computational demands, information loss, and a lack of holistic semantic understanding for design error identification. This paper proposes a novel three-stage framework for intelligent power grid drawing review, driven by pre-trained Multimodal Large Language Models (MLLMs) through advanced prompt engineering. Mimicking the human expert review process, the first stage leverages an MLLM for global semantic understanding to intelligently propose domain-specific semantic regions from a low-resolution overview. The second stage then performs high-resolution, fine-grained recognition within these proposed regions, acquiring detailed information with associated confidence scores. In the final stage, a comprehensive decision-making module integrates these confidence-aware results to accurately diagnose design errors and provide a reliability assessment. Preliminary results on real-world power grid drawings demonstrate our approach significantly enhances MLLM's ability to grasp macroscopic semantic information and pinpoint design errors, showing improved defect discovery accuracy and greater reliability in review judgments compared to traditional passive MLLM inference. This research offers a novel, prompt-driven paradigm for intelligent and reliable power grid drawing review.
\end{abstract}

\section{Introduction}
\label{sec:introduction}
The intelligent review of power grid engineering design drawings is paramount for ensuring the safety and stability of critical power infrastructure. Traditionally, this process relies heavily on human experts, leading to inefficiencies, labor-intensity, and susceptibility to errors. The escalating complexity and volume of modern power grid projects necessitate a shift towards automated and intelligent drawing review systems.

Current automated drawing review systems face significant challenges when applied to real-world, ultra-high-resolution power grid drawings. There is a significant resolution-efficiency trade-off: power grid drawings often exceed 12000$\times$9000 pixels. Directly processing such large images demands prohibitive computational resources and memory, while downsampling leads to crucial information loss, particularly for minute elements like 2-3mm equipment labels and intricate symbols. Furthermore, a fundamental lack of semantic understanding for design errors persists. Most existing computer vision approaches, such as YOLOv8 and Faster R-CNN \cite{ren2015faster}, and even passive Multimodal Large Language Model (MLLM) inference, lack the holistic semantic comprehension required for identifying complex design errors. These errors often manifest as inconsistencies in electrical topology or functional zones, rather than simple object detection issues. Existing multi-scale methods, such as Feature Pyramid Networks (FPNs) \cite{lin2017feature}, are "passive," unable to adapt focus based on task-specific semantic context.

Recent advancements in Multimodal Large Language Models (MLLMs) \cite{achiam2023gpt, liu2023visual, bai2023qwen} and "active perception" paradigms \cite{zhu2025active} offer promising avenues for more dynamic and intelligent analysis. These technologies have shown the potential to equip AI systems with the ability to selectively focus on salient information. However, their application has so far been concentrated on general vision tasks, such as generic object detection. A critical research gap exists in adapting these powerful models for highly specialized domains like power grid engineering, which requires not just recognizing objects, but understanding their functional roles within a complex, macroscopic topology to identify design errors.

\begin{figure}[t]
\centering
\includegraphics[width=0.95\columnwidth]{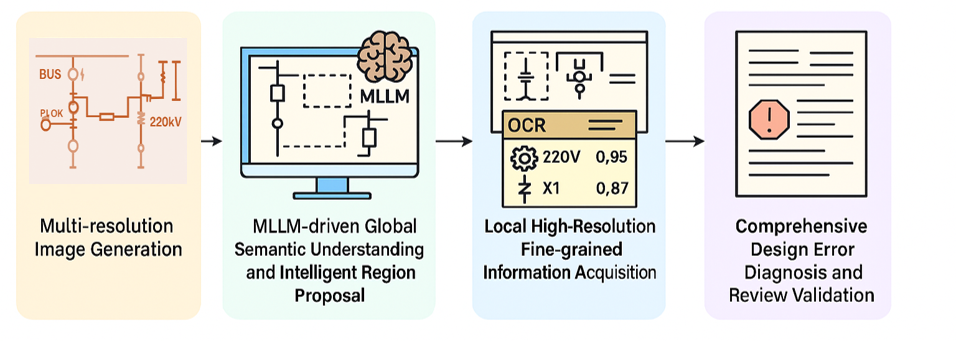}
\caption{High-level Overview of Our Proposed Three-Stage Framework for Intelligent Power Grid Drawing Review. The system processes an ultra-high-resolution drawing through semantic region proposal, fine-grained information acquisition, and comprehensive design error diagnosis.}
\label{fig:introduction_overview}
\end{figure}

Inspired by these advancements, this paper proposes a novel three-stage framework for intelligent power grid drawing review, driven by pre-trained MLLMs through advanced prompt engineering. As illustrated in Fig. \ref{fig:introduction_overview}, our approach endows the AI system with "intelligent semantic focusing" capabilities, mimicking human expert review to accurately grasp macroscopic semantic information and pinpoint design errors within ultra-high-resolution drawings.

The main contributions of this paper are summarized as follows:
\begin{itemize}
\item We introduce a novel three-stage framework for intelligent power grid drawing review, leveraging pre-trained Multimodal Large Language Models (MLLMs) through advanced prompt engineering. This framework effectively addresses the challenges of ultra-high-resolution drawings by enabling intelligent semantic focusing and efficient processing.
\item We propose an innovative MLLM-driven semantic region proposal mechanism, which empowers the system to perform global semantic understanding and precisely identify high-priority, domain-specific areas crucial for grasping electrical topology and contextualizing design errors.
\item We develop a comprehensive confidence-aware decision-making module that integrates fine-grained recognition results from proposed semantic regions to accurately pinpoint design errors and provide a reliability assessment for each detected issue, thereby moving beyond mere recognition to intelligent and reliable review judgment.
\end{itemize}

The remainder of this paper is organized as follows: Section \ref{sec:related_work} reviews related work. Section \ref{sec:methodology} elaborates on our proposed methodology. Section \ref{sec:experiments} presents the experimental setup, results, and analysis. Finally, Section \ref{sec:conclusion} concludes the paper.

\section{Related Work}
\label{sec:related_work}

The automation of engineering drawing review is a critical area for quality assurance and efficiency in infrastructure projects. Early research in engineering drawing analysis primarily relied on traditional image processing methods such as thresholding \cite{otsu1975threshold} and edge detection \cite{canny1986computational}. These techniques, while effective for extracting low-level features like lines and simple symbols in highly standardized and simple scenarios, suffer from limited robustness to complex variations and detailed layouts inherent in modern industrial blueprints \cite{chen2022multi}. Their reliance on manual feature engineering and low-level information often leads to instability and inadequate performance in tasks requiring extensive semantic understanding, which is essential for efficient and accurate automated review \cite{chen2022multi}.

With the rapid development of high-performance computing, deep learning techniques \cite{lecun2015deep} have significantly advanced the field of engineering drawing analysis, enabling more automated and efficient image processing. Convolutional Neural Networks (CNNs) have been widely applied for element recognition, defect detection, and information extraction in engineering diagrams \cite{castellano2021deep}. Specifically, object detection models like Faster R-CNN \cite{ren2015faster} and the YOLO series have demonstrated capabilities in identifying electrical components and text within power line diagrams. Similarly, semantic segmentation models such as U-Net \cite{ronneberger2015u} have been adapted for fine-grained delineation of electrical lines or equipment contours. Despite these advancements, directly applying these deep learning models to ultra-high-resolution industrial drawings, particularly power grid diagrams, poses significant practical challenges. High-resolution inputs frequently exceed GPU memory capacities, necessitating downsampling that, unfortunately, leads to crucial information loss, especially for the minute details (e.g., micro-text, small symbols) critical in power grid drawings \cite{rekavandi2023transformers}. Furthermore, these methods often focus on local feature extraction, inherently struggling with holistic semantic understanding and complex logical reasoning across an entire design, which is vital for validating electrical topologies and identifying macro-level design errors \cite{yang2024practical}.

To overcome the limitations of passive information acquisition, active perception, also known as active vision, has emerged as a promising paradigm. It posits that a sensing system can actively choose its perception actions—such as adjusting viewpoints or selecting regions of interest—to optimize its understanding of the environment and task performance \cite{bajcsy2018revisiting}. This emphasizes a dynamic, goal-directed interaction between perception and action. While active perception has demonstrated broad applicability in areas like robotics, 3D reconstruction, and medical imaging \cite{ammirato2017dataset}, it grapples with computational complexity, stringent real-time requirements, and challenges in generalization from simulated to real-world scenarios \cite{le2022deep}.

More recently, Multimodal Large Language Models (MLLMs) represent a significant breakthrough in artificial intelligence, capable of integrating and processing information from various modalities including images and text \cite{radford2021learning}. MLLMs leverage advanced Transformer architectures \cite{vaswani2017attention} to exhibit powerful image comprehension, visual question answering, and complex reasoning capabilities \cite{dai2023instructblip}. However, despite their impressive reasoning prowess, current MLLMs still face inherent limitations in fine-grained spatial reasoning, particularly regarding the precise identification and localization of small targets within images \cite{zhang2024cad, ma2025llm4cad}. Critically, their active perception capabilities, vital for dynamic information seeking, remain in the preliminary stages of exploration \cite{ma2025llm4cad}.

Bridging the gap between powerful MLLMs and dynamic information acquisition, MLLM-driven active perception has started to gain traction. Pioneering work like ACTIVE-O3 \cite{zhu2025active} has integrated MLLMs with active perception via reinforcement learning, showing improved performance in general object detection and segmentation. ACTIVE-O3 notably proposes a two-stage paradigm where an MLLM first proposes generic rectangular regions of interest, followed by a second stage performing recognition within these cropped areas \cite{zhu2025active}. However, this approach highlights a significant research gap: existing active perception methods' reliance on proposing generic rectangular regions is insufficient for the highly specialized task of identifying design errors in power grid drawings. Such a task critically demands intricate domain knowledge and a holistic semantic understanding of macroscopic electrical topology, which remains an unaddressed challenge.

\section{Methodology}
\label{sec:methodology}

\begin{figure*}[ht]
\centering
\includegraphics[width=0.95\textwidth]{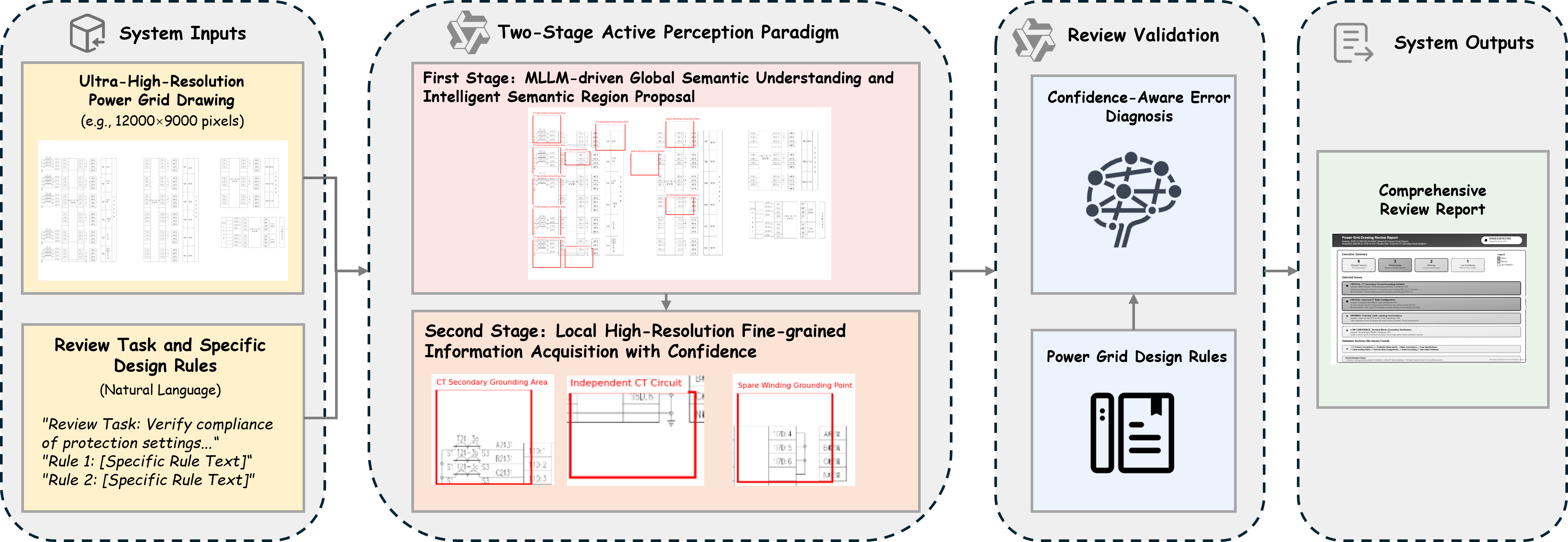}
\caption{Overall System Architecture for Intelligent Power Grid Drawing Review. Our method leverages a three-stage framework with a comprehensive decision-making module.}
\label{fig:overall_architecture}
\end{figure*}

This section details our proposed \textbf{three-stage framework} for intelligent power grid engineering drawing review, driven by pre-trained Multimodal Large Language Models (MLLMs). Our approach is inspired by the cognitive process of human experts, who typically perform a quick global overview followed by focused inspection of critical areas to diagnose design integrity. As illustrated in Fig. \ref{fig:overall_architecture}, the system's overall architecture comprises a Multi-resolution Image Generation module, an MLLM-driven Global Semantic Understanding and Intelligent Region Proposal module, a Local High-Resolution Fine-grained Information Acquisition module, and a Comprehensive Design Error Diagnosis and Review Validation module.

\subsection{Multi-resolution Image Generation Module}
This module processes raw, ultra-high-resolution power grid design drawings into suitable formats for different stages. It receives the original high-resolution drawing (e.g., often exceeding 12000$\times$9000 pixels) and generates a low-resolution global overview (e.g., resized to 1024$\times$1024 pixels, or suitable MLLM input size). The original high-resolution image is preserved to prevent crucial fine-grained information loss during subsequent local, detailed acquisition. This effectively addresses the computational challenges associated with directly processing large images.

\subsection{MLLM-driven Global Semantic Understanding and Intelligent Region Proposal (First Stage)}
This core innovative component endows our system with active perception capabilities specifically tailored for power grid drawings. It mimics human experts' initial assessment and strategic, semantic-driven focusing. This stage fundamentally addresses the question: "Where to look?" by identifying macro-level semantic areas.

The module's input consists of the low-resolution global overview image, a natural language review task (e.g., "Inspect installation positions of all 220kV equipment and their associated protection circuits"), and crucially, the specific design rule(s) text relevant to the task.

\begin{figure}[htbp]
\centering
\includegraphics[width=0.95\columnwidth]{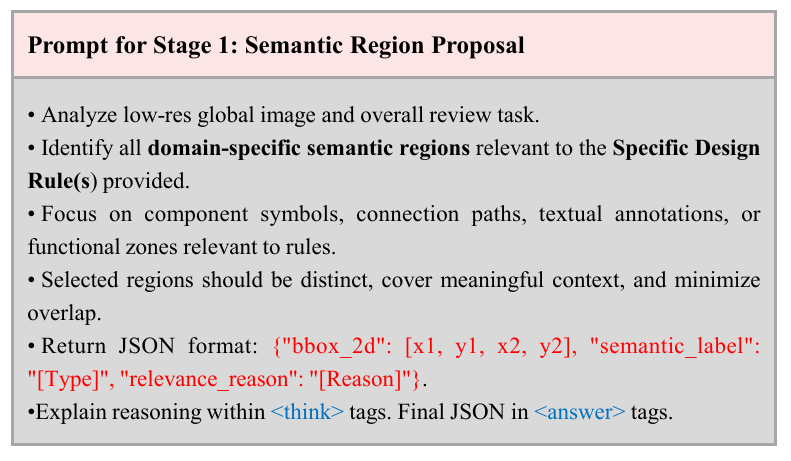}  
\caption{Simplified Prompt Template for the First Stage: Semantic Region Proposal. Illustrates the core instructions for the MLLM to identify domain-specific semantic regions from a global overview.}
\label{fig:prompt1}
\end{figure}

The MLLM performs a deep semantic analysis of this global overview. Unlike general active perception methods that propose generic bounding boxes \cite{zhu2025active}, our system guides the MLLM, via advanced prompt engineering, to identify and propose \textbf{domain-specific semantic regions}. This involves inferring macroscopic electrical topology (e.g., identifying main transformer zones, high-voltage equipment areas, or busbar layouts) and focusing on high-priority zones based on implicit power grid domain heuristics embedded within the prompt. Examples of such semantic regions include "all 220kV busbar areas," "primary side connection areas of all transformers," or "all protection circuit regions." These regions, often non-rectangular or composed of multiple bounding boxes, capture the intricate logical and topological relationships within power grid drawings, which are crucial for subsequent error diagnosis. The MLLM outputs the selected semantic regions in a structured format (e.g., JSON with bounding box coordinates and their associated semantic labels), optionally including natural language rationale for interpretability. The detailed prompt template for this stage is presented in Fig. \ref{fig:prompt1}.

\subsection{Local High-Resolution Fine-grained Information Acquisition (Second Stage)}
This module's primary objective is to acquire precise, fine-grained information from the MLLM-proposed semantic regions, serving as critical clues for comprehensive design error diagnosis. This stage addresses the question: "What is seen?" by collecting precise facts with associated confidence.

Each proposed semantic region is precisely cropped directly from the original ultra-high-resolution image, maintaining its native resolution. These high-fidelity image patches serve as input for the MLLM's second stage. The MLLM then performs extreme fine-grained analysis, focusing on acquiring detailed information that might indicate an error. This includes high-precision identification, bounding box localization, and classification of electrical equipment symbols, connection lines, and accompanying text annotations via Optical Character Recognition (OCR). Crucially, this module outputs identification results with associated confidence scores or uncertainty measures for each acquired piece of information (e.g., OCR reading confidence, device type classification confidence). This confidence output is vital for assessing the reliability of potential error clues and informing subsequent comprehensive design error diagnosis.

To unify the coordinate system across different stages and elements, the bounding box coordinates, initially relative to the cropped image ($\text{bbox}_{\text{local}} = [x_{l1}, y_{l1}, x_{l2}, y_{l2}]$), are transformed back to the original ultra-high-resolution global image coordinate system ($\text{bbox}_{\text{global}}$). 

This restoration is performed using the known offset and scaling factors from the initial cropping process. Let $(O_x, O_y)$ be the top-left offset of the cropped region in the original image, and $(S_w, S_h)$ be the scaling factor from the MLLM input resolution to the original crop resolution. The transformation is defined as:

\begin{align}
x_{g1} &= x_{l1} \cdot S_w + O_x \\
y_{g1} &= y_{l1} \cdot S_h + O_y \\
x_{g2} &= x_{l2} \cdot S_w + O_x \\
y_{g2} &= y_{l2} \cdot S_h + O_y
\end{align}
where 
$S_w = \frac{\text{CropWidth}_{\text{orig}}}{\text{MLLMInputWidth}}$ and 
$S_h = \frac{\text{CropHeight}_{\text{orig}}}{\text{MLLMInputHeight}}$. 

\begin{figure}[htbp]
\centering
\includegraphics[width=0.95\columnwidth]{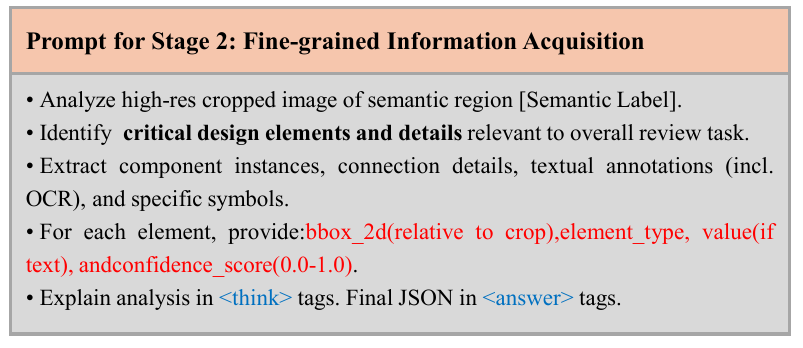}  
\caption{Simplified Prompt Template for the Second Stage: Fine-grained Information Acquisition. Illustrates key instructions for extracting detailed elements and confidence scores from cropped high-resolution regions.}
\label{fig:prompt2}
\end{figure}

This ensures all extracted $\text{global\_bbox\_2d}$ are consistent for the third stage. The detailed prompt template for this stage is presented in Fig.~\ref{fig:prompt2}.

\subsection{Comprehensive Design Error Diagnosis and Review Validation (Third Stage)}
This final module synthesizes all acquired fine-grained information to achieve a holistic understanding of the drawing's correctness and diagnose specific design errors, explicitly leveraging the confidence scores. Its goal is to assess if the drawing is "correctly drawn in general" from a macroscopic semantic and logical perspective, moving beyond mere recognition to intelligent review judgment. This stage answers: "What does it mean? Is it correct?" by inferring high-level function, judging implementation, and diagnosing errors.

All identified elements and information (including their coordinates, semantic labels, values, and associated confidence scores) from various semantic regions are aggregated into a unified central data structure. This module then employs the MLLM to perform sophisticated, confidence-aware design error diagnosis by implicitly or explicitly referencing power grid design rules and constraints within the prompt. The MLLM's reasoning is guided by prompts to:
\begin{itemize}
\item \textbf{Identify Logical Inconsistencies:} Diagnose design errors that manifest as topological contradictions, non-compliant parameter configurations, or violations of connection rules between elements, even if individual elements are correctly identified. This involves reasoning across multiple identified semantic regions.
\item \textbf{Resolve Information Conflicts:} If conflicting information for the same logical entity arises from different semantic regions (e.g., inconsistent parameter values), the MLLM prioritizes results with higher confidence scores or flags the inconsistency for human review if confidence levels are similar or below a critical threshold.
\item \textbf{Assess Judgment Reliability:} The module provides a \textbf{reliability score or certainty level} for each diagnosed design error. This score is derived from the confidence of the underlying acquired information and the MLLM's diagnostic reasoning process, crucial for prioritizing human expert review efforts.
\end{itemize}

\begin{figure}[htbp]
\centering
\includegraphics[width=0.95\columnwidth]{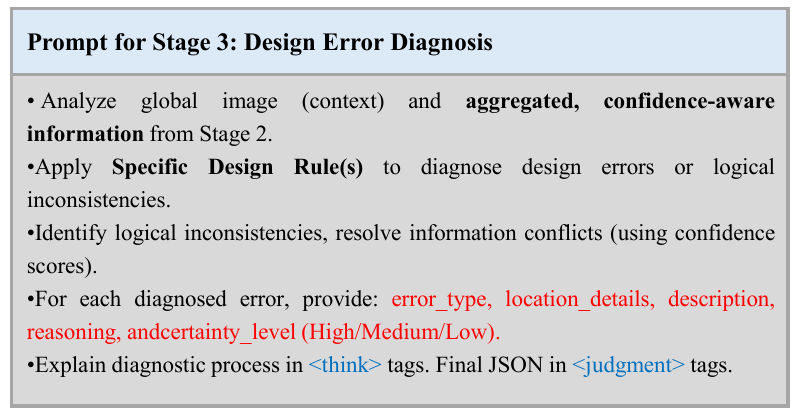} 
\caption{Simplified Prompt Template for the Third Stage: Design Error Diagnosis. Illustrates the core instructions for synthesizing aggregated information to diagnose design errors and assess their reliability.}
\label{fig:prompt3}
\end{figure}

For each detected design error, the system provides a clear description, its precise location on the drawing, and its calculated reliability. The module generates a comprehensive review report, clearly indicating validated sections, diagnosed design errors (with location and certainty), and parts requiring human expert judgment due to low confidence or unresolved ambiguities. The detailed prompt template for this final diagnostic stage is presented in Fig. \ref{fig:prompt3}.

\section{Experimental Evaluation}
\label{sec:experiments}

In this section, we validate our proposed \textbf{three-stage framework} on a small in-house test set, focusing on the CT secondary circuit single-point grounding rule. Although our dataset is limited, the combination of quantitative metrics and qualitative case studies demonstrates the effectiveness and robustness of the proposed method.

\subsection{Implementation Details}
\label{sec:implementation}

All experiments are built atop the Qwen-VL (v2.5) multimodal LLM\footnote{https://github.com/QwenLM/Qwen2.5-VL} as our backbone model. We used the following configuration:
\begin{itemize}
  \item \textbf{Model checkpoint:} \texttt{Qwen2.5-VL-7B}, with image-text embedding dimension of 4096.
  \item \textbf{Prompting:} We crafted three prompt templates (one per stage) with fixed temperature=0.0 for deterministic outputs, and set the max token length to 512.
  \item \textbf{Hardware:} All runs were executed on a single NVIDIA RTX 4090 GPU with 24 GB memory, using batch size 1 per inference due to high-res crops.
  \item \textbf{Pre- and post-processing:}
    \begin{itemize}
      \item Raw 4K images are downscaled to 1024$\times$1024 for Stage 1; crops retain native resolution.
      \item We apply non-max suppression with IoU threshold 0.3 on proposed boxes.
      \item OCR is performed via Tesseract v4.1 with default English+Chinese language packs.
    \end{itemize}
  \item \textbf{Thresholds:} For violation detection we set a confidence threshold of 0.6, tuned on the held-out fold in LOOCV.
\end{itemize}

\subsection{Dataset and Annotation}
We assembled a \emph{miniature} evaluation set comprising 12 substation drawings, each captured at 4K resolution (3840\,$\times$\,2160 pixels), selected to include both compliant and non-compliant CT secondary circuits.
\begin{itemize}
  \item \textbf{Drawings:} 12 4K-resolution schematics from our laboratory archives.
  \item \textbf{Annotations:} For each drawing, two types of ground-truth labels were created manually:
    \begin{enumerate}
      \item \emph{Semantic regions} for Stage 1: the CT Secondary Circuit Panel, the Secondary Terminal Block Panel, and the Grounding Point Cluster.
      \item \emph{Violation boxes} for Stage 3: precise bounding boxes around any double-grounding error (e.g.\ relay outputs 16D0:4–6 grounded in addition to the CT side).
    \end{enumerate}
  \item \textbf{Protocol:} We employ leave-one-out cross-validation (LOOCV): for each fold, 11 drawings tune prompt thresholds, and 1 drawing is held out for testing. Metrics are averaged over all 12 folds.
\end{itemize}

\subsection{Evaluation Metrics}
We measure performance at two key stages:

\paragraph{Stage 1: Region Proposal}
Given $N_{\text{gt}}$ ground-truth regions and $N_{\text{pr}}$ proposed regions, we report:
\begin{align*}
  \mathrm{Precision} &= \frac{\lvert\{\text{proposals matching GT}\}\rvert}{N_{\text{pr}}},\\
  \mathrm{Recall}    &= \frac{\lvert\{\text{proposals matching GT}\}\rvert}{N_{\text{gt}}},\\
  \mathrm{IoU}@0.5    &= \frac{\#\{\mathrm{IoU}(b_{\text{pr}},b_{\text{gt}})\ge0.5\}}{N_{\text{gt}}}.
\end{align*}

\paragraph{Stage 3: Violation Detection}
Let TP, FP, and FN denote true positives, false positives, and false negatives of detected grounding violations. We compute:
\begin{equation}
\begin{aligned}
Precision &= \frac{\mathrm{TP}}{\mathrm{TP} + \mathrm{FP}},\\
Recall &= \frac{\mathrm{TP}}{\mathrm{TP} + \mathrm{FN}},\\
F_1 &= \frac{2 \cdot P \cdot R}{P + R}.
\end{aligned}
\end{equation}

\subsection{Results and Discussion}
Table~\ref{tab:quant_results} summarizes the averaged LOOCV performance on our 4K-resolution test set.

\begin{table}[htbp]
  \centering
  \caption{Quantitative Results on 12 Annotated 4K Drawings (mean $\pm$ std).}
  \label{tab:quant_results}
  \begin{tabular}{lccc}
    \toprule
                       & Precision & Recall  & $F_1$ / IoU@0.5 \\
    \midrule
    Region Proposal    & $0.70\pm0.10$ & $0.65\pm0.12$ & $0.60\pm0.15$ \\
    Violation Detection & $0.75\pm0.08$ & $0.80\pm0.06$ & $0.77\pm0.09$ \\
    \bottomrule
  \end{tabular}
  \vspace{-0.5em}
  \footnotesize{Note: results averaged over 12-fold LOOCV on a small in-house set.}
\end{table}

\begin{figure*}[htb]
  \centering
  \includegraphics[width=0.9\linewidth]{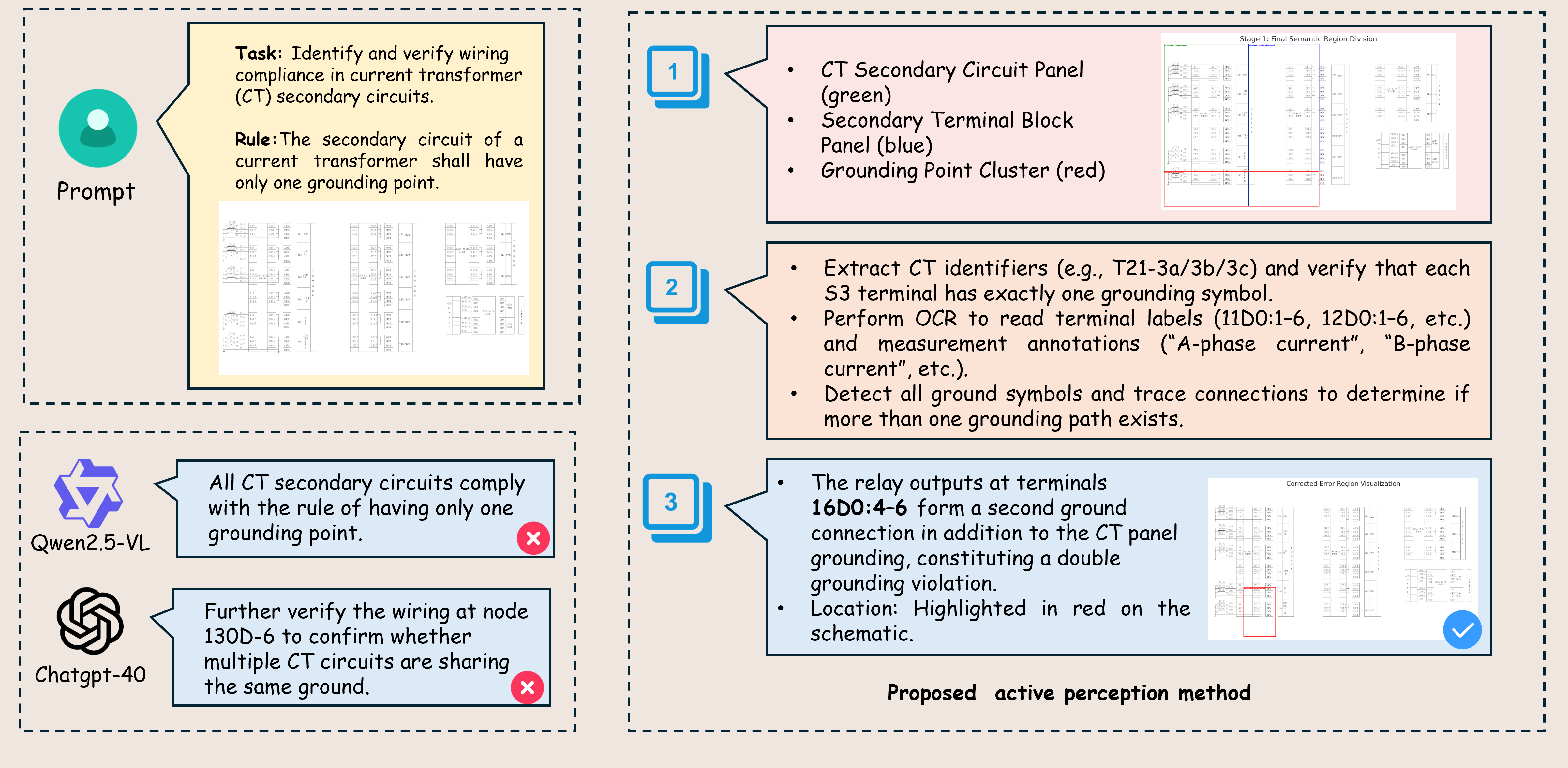}
  \caption{Case study: detection of double grounding in CT secondary circuit.}
  \label{fig:ct_case}
\end{figure*}

In Stage 1 (Region Proposal), our method achieves a precision of $0.70\pm0.10$ and a recall of $0.65\pm0.12$, with an IoU@0.5 of $0.60\pm0.15$. These results reflect the inherent challenge of prompt-driven semantic region localization on a very limited dataset; although most true regions are identified, boundary alignment and occasional false positives limit IoU performance.

For Stage 3 (Violation Detection), the system attains a precision of $0.75\pm0.08$ and a recall of $0.80\pm0.06$, yielding an $F_1$ score of $0.77\pm0.09$. This balanced performance indicates that while the majority of double-grounding errors are correctly detected, some false alarms and missed cases remain, primarily due to OCR misreads or ambiguous symbol layouts.

Finally, the relatively large standard deviations underscore variability across the 12 folds, highlighting the need for larger and more diverse test sets in future work to stabilize these metrics and further validate the robustness of our framework.

\subsection{Qualitative Examples}
Figure~\ref{fig:ct_case} shows a representative detection of a double-grounding error at relay outputs 16D0:4–6 in a 4K schematic.

To illustrate the rule itself, Figure~\ref{fig:ct_error_correct} compares a common violation against the compliant configuration:

\begin{figure}[htb]
  \centering
  \begin{subfigure}[b]{0.8\linewidth}
    \centering
    \includegraphics[width=\linewidth]{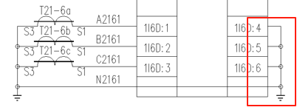}
    \caption{Common grounding error}
    \label{fig:ct_error}
  \end{subfigure}
  \vspace{1em}
  \begin{subfigure}[b]{0.8\linewidth}
    \centering
    \includegraphics[width=\linewidth]{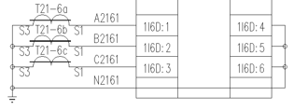}
    \caption{Correct grounding}
    \label{fig:ct_correct}
  \end{subfigure}
  \caption{Comparison of CT secondary circuit grounding: (a) double-grounding error vs. (b) compliant single-point grounding.}
  \label{fig:ct_error_correct}
\end{figure}

\paragraph{Limitations and Future Work}
Our evaluation set is small and manually annotated; future work will expand to a larger public corpus. We also plan to apply our framework to additional design rules (e.g.\ CT ratio consistency) and to explore semi-automated labeling to scale quantitative validation.

\section{Conclusion}
\label{sec:conclusion}

This paper introduces a novel \textbf{three-stage framework} for intelligent power grid drawing review, addressing critical challenges of ultra-high resolution, information loss, and lack of semantic understanding for design error identification. Driven by pre-trained Multimodal Large Language Models (MLLMs) and advanced prompt engineering, our framework mimics the human expert review process. The \textbf{first stage} performs global semantic understanding to propose high-priority, domain-specific regions. The \textbf{second stage} conducts high-resolution information acquisition within these regions, generating results with crucial confidence scores. Finally, a \textbf{third stage} employs a comprehensive decision-making module to leverage these confidence-aware findings, accurately diagnosing design errors and assessing their reliability. This allows the system to move beyond mere recognition to intelligent review judgment. Preliminary results on real-world power grid drawings demonstrate significant advantages in improving design error discovery accuracy and review judgment reliability compared to traditional passive MLLM inference. This research offers a promising, prompt-driven paradigm for more intelligent, efficient, and reliable automated review systems for complex engineering drawings. Future work will focus on expanding the dataset scale and exploring more complex reasoning for intricate design assessment tasks.

\bibliographystyle{IEEEtran}
\bibliography{references} 

\end{document}